\documentclass[
]{ceurart}

\usepackage{listings}
\begin{document}

\copyrightyear{2023}
\copyrightclause{Copyright for this paper by its authors.
  Use permitted under Creative Commons License Attribution 4.0
  International (CC BY 4.0).}

\conference{In A. Martin, K. Hinkelmann, H.-G. Fill, A. Gerber, D. Lenat, R. Stolle, F. van Harmelen (Eds.), 
Proceedings of the AAAI 2023 Spring Symposium on Challenges Requiring the Combination of Machine Learning and Knowledge Engineering (AAAI-MAKE 2023), Hyatt Regency, San Francisco Airport, California, USA, March 27-29, 2023.}

\title{An Independent Evaluation of ChatGPT on Mathematical Word Problems (MWP)}

\author{Paulo Shakarian}[%
orcid=,
email=pshak02@asu.edu,
url=https://labs.engineering.asu.edu/labv2/,
]
\cormark[1]
\address{Arizona State University,
  699 S Mill Ave, Tempe, AZ, 85281, USA}
\author{Abhinav Koyyalamudi}[%
orcid=,
email=akoyyala@asu.edu,
url=,
]

\author{Noel Ngu}[%
orcid=,
email=nngu2@asu.edu,
url=,
]

\author{Lakshmivihari Mareedu}[%
orcid=,
email=lmareedu@asu.edu,
url=,
]

\cortext[1]{Corresponding author.}



\begin{abstract}
We study the performance of a commercially available large language model (LLM) known as ChatGPT on math word problems (MWPs) from the dataset DRAW-1K.  To our knowledge, this is the first independent evaluation of ChatGPT.  We found that ChatGPT's performance changes dramatically based on the requirement to show its work, failing $20\%$ of the time when it provides work compared with $84\%$ when it does not.  Further several factors about MWPs relating to the number of unknowns and number of operations that lead to a higher probability of failure when compared with the prior, specifically noting (across all experiments) that the probability of failure increases linearly with the number of addition and subtraction operations.  We also have released the dataset of ChatGPT's responses to the MWPs to support further work on the characterization of LLM performance and present baseline machine learning models to predict if ChatGPT can correctly answer an MWP.  We have released a dataset comprised of ChatGPT's responses to support further research in this area.
\end{abstract}


\maketitle

\section{Introduction}
The emergence of large language models (LLM) has gained much popularity in recent years.  At the time of this writing, some consider OpenAI's GPT 3.5 series models as the state-of-the art~\cite{gptAbility}.  In particular, a variant tuned for natural dialogue known as ChatGPT~\cite{chatGptBlog}, released in November 2022 by OpenAI, has gathered much popular interest, gaining over one million users in a single week~\cite{cgptUsers}.  However, in terms of accuracy, LLMs are known to have performance issues, specifically when reasoning tasks are involved~\cite{gptAbility,hoffmann_training_2022}.  This issue, combined with the ubiquity of such models has led to work on prompt generation and other aspects of the input~\cite{wei_chain--thought_nodate,wang_self-consistency_2022}.  Other areas of machine learning, such as meta-learning~\cite{hospedales_meta-learning_2022,zhou_domain_2022} and introspection~\cite{daftry_introspective_2016,ramanagopal_failing_2018} attempt to predict when a model will succeed or fail for a given input.  An introspective tool, especially for certain tasks, could serve as a front-end to an LLM in a given application.

As a step toward such a tool, we investigate aspects of math word problems (MWPs) that can indicate the success or failure of ChatGPT on such problems.  We found that ChatGPT's performance changes dramatically based on the requirement to show its work, failing $20\%$ of the time when it provides work compared with $84\%$ when it does not.  Further several factors about MWPs can lead to a higher probability of failure when compared with the prior, specifically noting that the probability of failure increases linearly with the number of addition and subtraction operations (across all experiments).  We also have released the dataset of ChatGPT's responses to the MWPs to support further work on the characterization of LLM performance.  While there has been previous work examining the LLM performance on MWPs~\cite{hoffmann_training_2022}, such work did not investigate specific aspects that increase MWP difficulty nor did it examine performance on ChatGPT in particular.

The remainder of this paper proceeds as follows.  In Section~\ref{sec:methSec}, we describe our methodology.  Then we describe our results in Section~\ref{sec:res}.  Using these intuitions, we present baseline models to predict the performance of ChatGPT in Section~\ref{sec:prefPred}.  This is followed by a discussion of related work (Section~\ref{sec:rw}) and future work (Section~\ref{sec:lim}).

\section{Methodology}
\label{sec:methSec}
\begin{figure}[h]
\centering
\includegraphics[scale=.5]{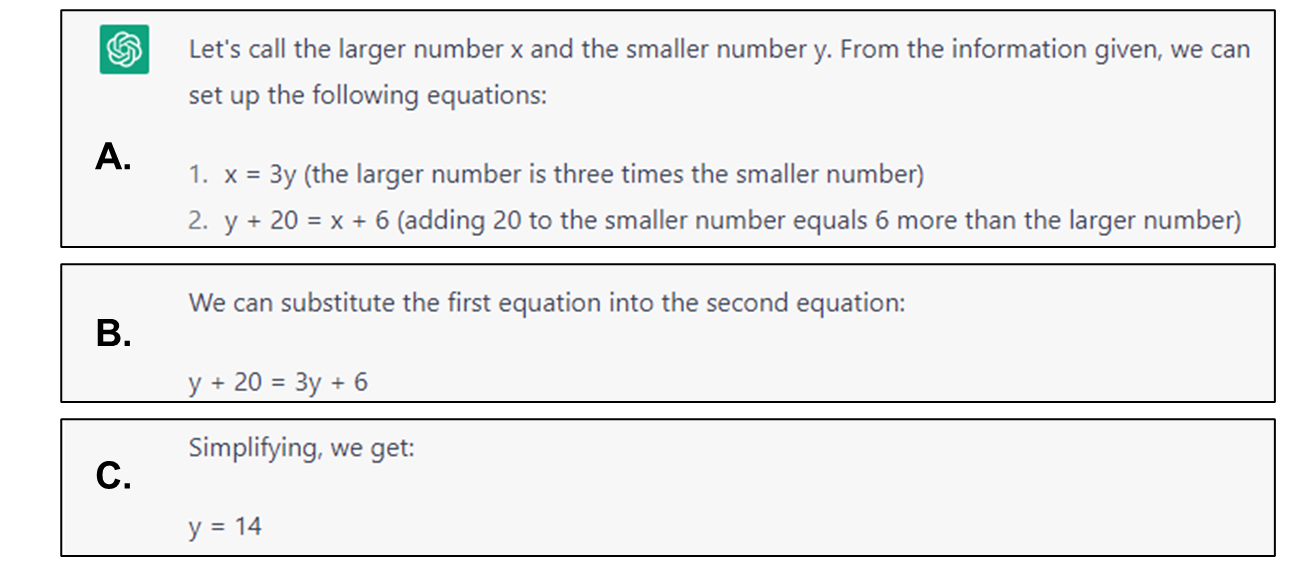}
\caption{ChatGPT's response (Jan. 24, 2023) to MWP \textit{One whole number is three times a second.  If 20 is added to the smaller number, the result is 6 more than the larger.}  In Step A it correctly identifies the set of equations needed to solve the problem and correctly simplifies it in Step B.  However, it fails to correctly perform the algebraic operation in Step C (it should state $2y=14$).  This leads ChatGPT to obtain an incorrect result, returning $42$ and $14$ instead of $21$ and $7$.}
\label{fig:exChat}
\end{figure}

\noindent\textbf{MWP Dataset.}  In our study, we employed the DRAW-1K dataset~\cite{upadhyay_learning_2016,upadhyay_annotating_2017,lan_mwptoolkit_2022} which not only includes 1,000 MWPs with associated answers but also template algebraic equations that one would use to solve such a word problem.  As a running example, consider the following MWP.
\begin{quote}
    \textit{One whole number is three times a second.  If 20 is added to the smaller number, the result is 6 more than the larger.}
\end{quote}  
We show ChatGPT's (incorrect) response to this MWP in Figure~\ref{fig:exChat}.  The DRAW-1K dataset not only includes the correct answer, which in this case is $12$ and $7$ but also includes template equations used to solve the problem.  For our running example, this consists of the equations $m - n = a - b$ and $c \times m - n = 0$. 
 This information represents a symbolic representation of the problem which can potentially be used to identify aspects that make such problems more difficult.
\vspace{10pt}

\noindent\textbf{Entering Problems into ChatGPT at Scale.} At the time of our study, OpenAI, the maker of ChatGPT had not released an API.  However, using the ChatGPT CLI Python Wrapper\footnote{We used ChatGPT CLI Python Wrapper by Mahmoud Mabrouk, see \url{https://github.com/mmabrouk/chatgpt-wrapper}} we interfaced with ChatGPT allowing us to enter the MWP's at scale.  For the first two experiments, we would add additional phrases to force ChatGPT to show only the final answer.  We developed these additions to the prompt based on queries to ChatGPT to generate the most appropriate phrase.  However, we found in our third experiment that this addition impacted results.  We ran multiple experiments to test ChatGPT's ability with these problems.
\begin{itemize}
\item \textbf{January 2023 Experiment (No work).}  Our first experiment was run in early January 2023 prior to OpenAI's announcement of improved performance on mathematical tasks  on January 30, 2023\footnote{\url{https://help.openai.com/en/articles/6825453-chatgpt-release-notes}} and in this experiment we included the following statement as part of the prompt.
\begin{quote}
\textsf{Don't provide any work/explanation or any extra text. Just provide the final number of answers for the previous question, with absolutely no other text. if there are two or more answers provide them as a comma separated list of numbers.}
\end{quote}
\item \textbf{February 2023 Experiment (No work).} Our second experiment was run in mid-February 2023 after the aforementioned OpenAI announcement and also used a prompt that would cause ChatGPT to show only the answer, however we found that our original prompt led to more erratic behavior, so we modified the prompt for this experiment, and used the following.
\begin{quote}
   \textsf{Don't provide any work/explanation or any extra text. Just provide the final number of answers for the previous question, with absolutely no other text. if there are two or more answers provide them as a comma separated list of numbers like: '10, 3,' etc; or if there is only 1 answer provide it like '10'. Absolutely no other text just numbers alone. Just give me the numbers (one or more) alone. No full stops, no spaces, no words, no slashes, absolutely nothing extra except the 1 or more numbers you might have gotten as answers.}
\end{quote}
\item \textbf{February 2023 Experiment (Showing Work).} We also repeated the February experiment without the additional prompt, thereby allowing ChatGPT to show all its work.  We note that in this experiment we used ChatGPT Plus which allowed for faster response.  At the time of this writing, ChatGPT Plus is only thought to be an improvement to accessibility and not a different model.\footnote{https://openai.com/blog/chatgpt-plus/}
\end{itemize}

\section{Results}
\label{sec:res}
\begin{figure}[h]
\centering
\includegraphics[scale=.3]{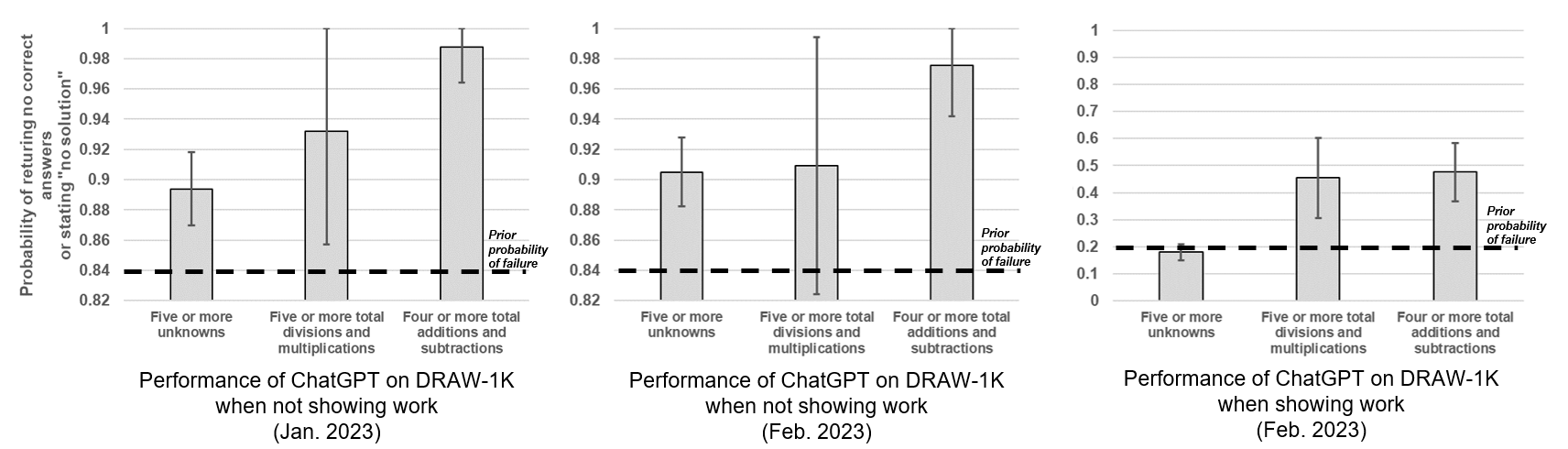}
\caption{Aspects of MWPs that led to ChatGPT failure more often than the prior ($95\%$ confidence intervals shown).}
\label{fig:gphs}
\end{figure}

\begin{figure}[h]
\centering
\includegraphics[scale=.38]{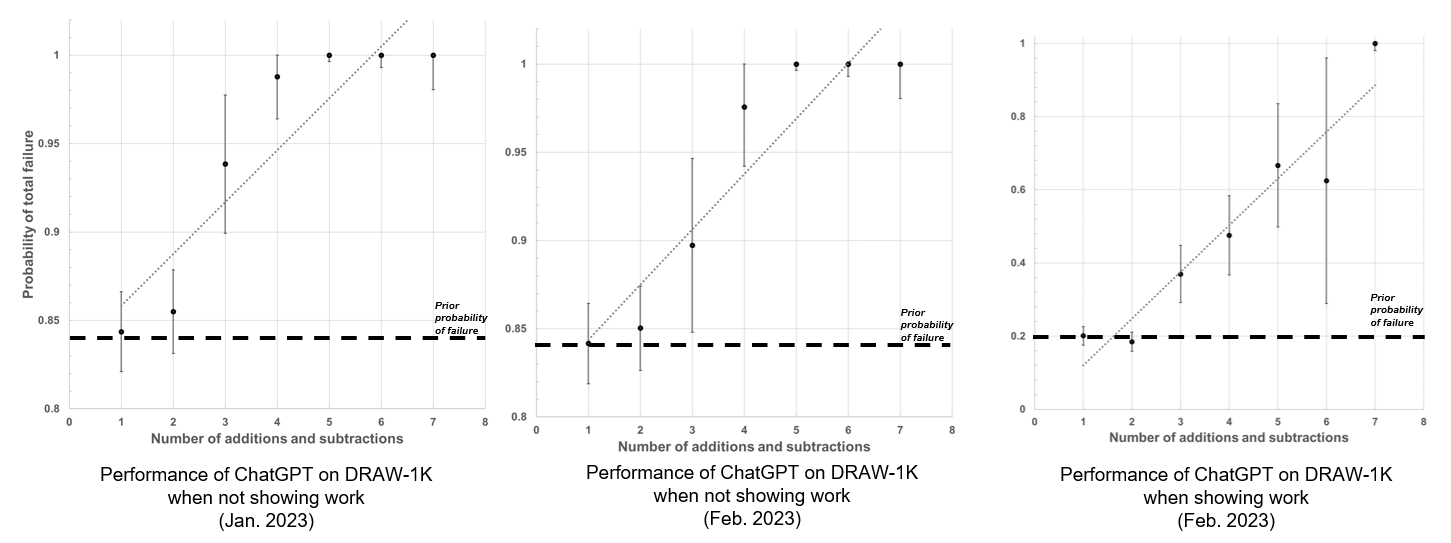}
\caption{Increase in probability of an incorrect response as a function of the number of addition operations (prior probability shown with dashed line, $95\%$ confidence intervals, linear regression with $R^2=0.821$ for January, $R^2=0.870$ for February without showing work and $R^2=0.915$ for February with showing work).}
\label{fig:regres}
\end{figure}

The key results of this paper are as follows: (1.) the creation of a dataset consisting of ChatGPT responses to the MWPs, (2.) identification of ChatGPT failure rates ($84\%$ for January and February experiments with no work and $20\%$ for the February experiment with work), (3.) identification of  several factors about MWPs relating to the number of unknowns and number of operations that lead to a higher probability of failure when compared with the prior (Figure~\ref{fig:gphs}), (4.) identification that the probability of failure increases linearly with the number of addition and subtraction operations (Figure~\ref{fig:regres}), and (5.) identification of a strong linear relationship between the number of multiplication and division operations and the probability of failure in the case where ChatGPT shows its work.  
\vspace{10pt}

\noindent\textbf{Dataset.}  We have released ChatGPT's responses to the 1,000 DRAW-1K MWP's for general use at \textbf{\url{https://github.com/lab-v2/ChatGPT_MWP_eval}}.  We believe that researchers studying this dataset can work to develop models that can combine variables, operate directly on the symbolic template, or even identify aspects of the template from the problem itself in order to predict LLM performance.  We note that at the time of this writing, collecting data at scale from ChatGPT is a barrier to such work as API's are not currently directly accessible, so this dataset can facilitate such ongoing research without the overhead of data collection.
\vspace{10pt}

\begin{figure}[h]
\centering
\includegraphics[scale=.4]{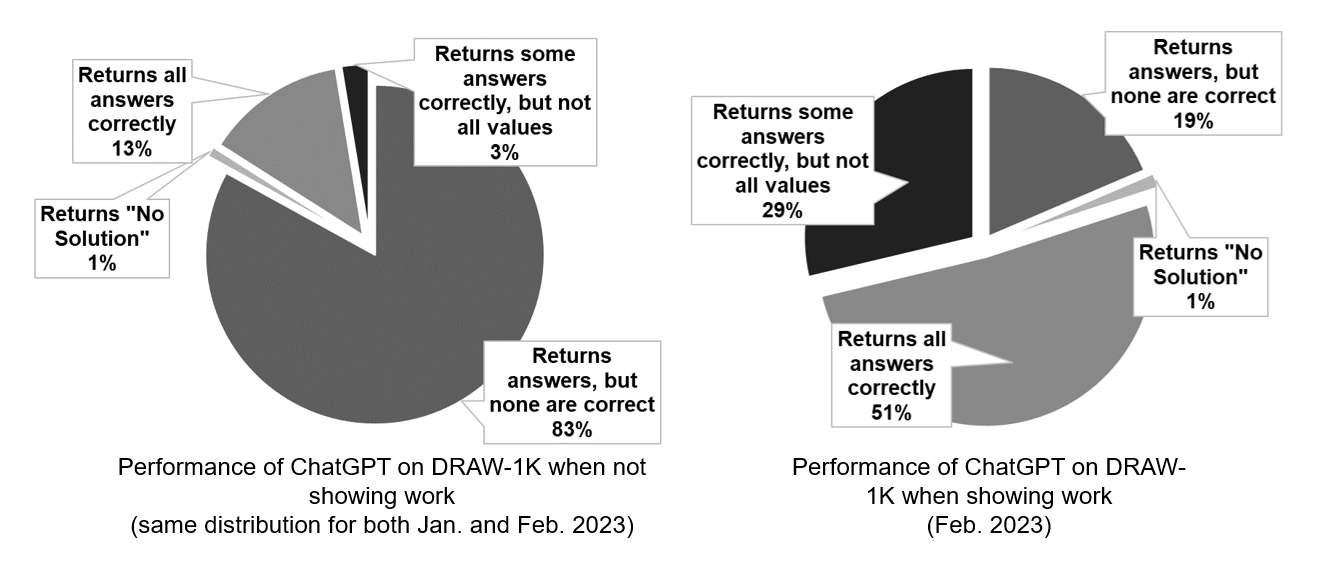}
\caption{Overall results on the 1,000 MWPs in DRAW-1K based on ChatGPT's response.}
\label{fig:overall}
\end{figure}

\noindent\textbf{Overall Performance of ChatGPT on DRAW-1K.}  As DRAW-1K provides precise can complete answers for each problem, we classified ChatGPT responses in several different ways and the percentage of responses in each case is shown in Figure~\ref{fig:overall}.
\begin{enumerate}
    \item \textit{Returns all answers correctly.}  Here ChatGPT returned all answers to the MWP (though it may round sometimes).
    \item \textit{Returns some answers correctly, but not all values.}  Here the MWP called for more than one value, but ChatGPT only returned some of those values.
    \item \textit{Returns ``No Solution.''} Here ChatGPT claims there was no solution to the problem.  This was not true for any of the problems.
    \item \textit{Returns answers, but none are correct.}  Here ChatGPT returned no correct answers (e.g., see Figure~\ref{fig:exChat}).
\end{enumerate}
Throughout this paper, we shall refer to the probability of failure as the probability of cases 3 and 4 above (considered together).  In our February experiment, we found that when ChatGPT omitted work, the percentages, as reported in Figure~\ref{fig:overall} remained the same, though they differed significantly when work was included.  We also report actual numbers for all experiments in Table~\ref{table:restable}.  We note that the probability of failure increases significantly when the work is not shown.  However, when the work is included, ChatGPT obtains performance in line with state-of-the-art models (i.e. EPT~\cite{kim_point_2020,kim_ept-x_2022}) which has a reported $59\%$ accuracy while ChatGPT (when work is shown) has fully correct (or rounded) answers $51\%$ of the time, but can be viewed as high as $80\%$ if partially correct answers are included. 

\begin{table}
\begin{center}
\begin{tabular}{ |c|c|c|c| }
 \hline
 Response Type & Jan. 2023  & Feb. 2023 & Feb. 2023\\
 & (No work)  & (No work) & (Showing work)\\

\hline
\hline
 Returns answers, but none are correct & 831 & 830 & 186 \\ 
 Returns ``No Solution''  & 9 & 10 & 14\\ 
Returns all answers correctly & 135 & 134 & 513 \\ 
Returns some answers correctly, but not all values & 25 & 26 & 287 \\
 \hline
\end{tabular}
\end{center}
\caption{Number of responses for each ChatGPT Variant}
\label{table:restable}
\end{table}

\noindent\textbf{Factors Leading to Incorrect Responses.}  We studied various factors from the templated solutions provided for the MWP in the DRAW-1K dataset and these included number of equations, number of unknowns, number of division and multiplication operations, number of addition and subtraction operations, and other variants derived from the metadata in the DRAW-1K dataset.  We identified several factors that, when present, cause ChatGPT to fail with a probability greater than the prior (when considering the lower bound of a $95\%$ confidence interval).  These results are shown in Figure~\ref{fig:gphs}.  One interesting aspect we noticed is that when the system would be required to show its work, the number of unknowns present no longer seems to increase the probability of failure (this was true for all quantities of unknowns in addition to what is shown in Figure~\ref{fig:gphs}).  Additionally, the number of multiplication and division operations, while increasing the probability of failure greater than the prior in the January experiment was not significant (based on $95\%$ confidence intervals) in the February experiment (when work was not shown) - possibly a result of OpenAI's improvements made at the end of January.  However, there was a significant relationship between the number of multiplication and division operations and failure when work was shown.  In fact, we found a strong linear relationship ($R^2=0.802$) for this relationship in the case where work was shown.

\begin{figure}[h]
\centering
\includegraphics[scale=.23]{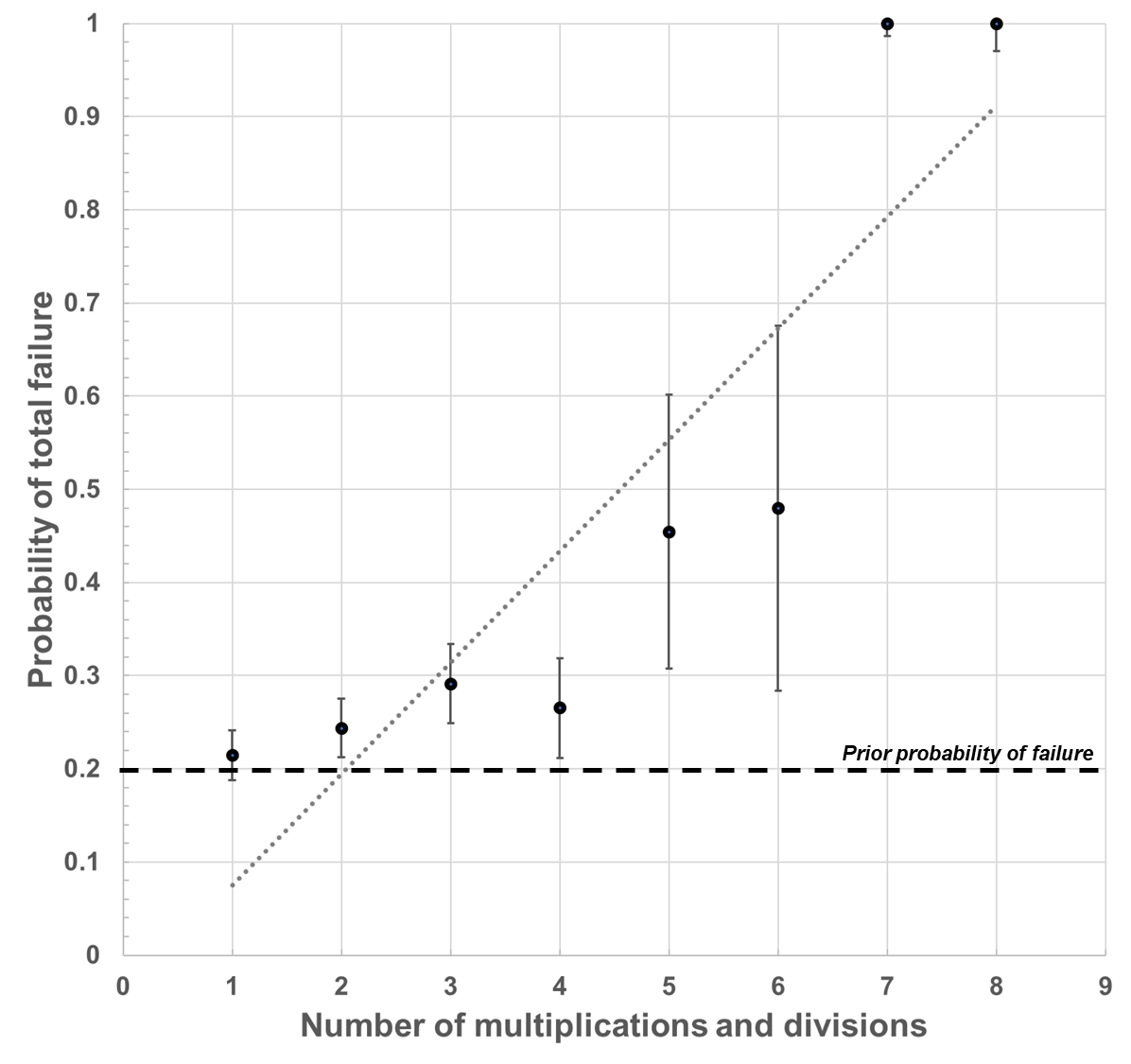}
\caption{Additional finding specific to the February, 2023 experiment where ChatGPT displayed its work relating number of multiplications to probability of failure,  $R^2=0.802$, $95\%$ confidence intervals.}
\label{fig:regresPlus}
\end{figure}

\noindent\textbf{Correlation of failure with additions and subtractions.}  Previous work has remarked on the failure of LLM's in multi-step reasoning~\cite{gptAbility,hoffmann_training_2022}.  In our study, we identified evidence of this phenomenon.  Specifically, we found a strong linear relationship between the number of addition and subtraction operations with the probability of failure ($R^2=0.821$ for the January experiment, $R^2=0.870$ for the February experiment and $R^2=0.915$ when work was shown).  We show this result in Figure~\ref{fig:regres}.  It is noteworthy that the relationship existed in all of our experiments, and seemed to be strengthened when ChatGPT included work in the result.

\section{Performance Prediction Baselines}
\label{sec:prefPred}
The results of the previous section, in particular, the factors indicating a greater probability of failure (e.g. Figures~\ref{fig:gphs}-\ref{fig:regres}), may indicate that the performance of ChatGPT can be predicted.  In this section, we use features obtained from the equations associated with the MWPs to predict performance.  Note that here we use ground-truth equations to derive the features, so the models presented in this section are essentially using an oracle - we leave extracting such features from equations returned by ChatGPT or another tool (e.g., EPT~\cite{kim_point_2020}) to future work.  That said, as these features deal with counts of operations, unknowns, and equations, a high degree of accuracy in creating the equations would not be required to faithfully generate such features.

Following the ideas of machine learning introspection~\cite{daftry_introspective_2016,ramanagopal_failing_2018}, we created performance prediction models using random forest and XGBoost.  We utilized scikit-learn 1.0.2 and XGBoost 1.6.2 respectively.  In our experiments, we evaluated each model on each dataset  using a five-fold cross-validation and report average precision and recall in Table~\ref{table:mltable} (along with F1 computed based on those averages).  In general, our models were able to provide higher precision than random on predicting incorrect answers for both classifiers.  Further, XGBoost was shown to be able to provide high recall for predicting correct responses.  While these results are likely not suitable for practical use, they do demonstrate that the features extracted provide some amount of signal to predict performance and provide a baseline for further study.

\begin{table}
\begin{center}
\begin{tabular}{ |c|c|c c c|c c c| }
 \hline
 Version of  & Model & Incorr. & Incorr. & Incorr. & Corr. & Corr. &Corr. \\
  ChatGPT &Type & Prec. & Recall & F1 & Prec. & Recall & F1 \\
 \hline
\hline
Jan. & RF & 0.90 & 0.88 & 0.89 & 0.34 & 0.41 & 0.37 \\
(No work)& XGBoost & 0.95 & 0.22 & 0.36 & 0.16 & 0.93 & 0.26\\
 \hline
Feb. & RF & 0.94 & 0.89 & 0.91 & 0.47 & 0.63 & 0.54\\
 (No work) & XGBoost & 0.98 & 0.35 & 0.51 & 0.18 & 0.95 & 0.31\\ 
  \hline
  Feb. & RF & 0.78 & 0.69 & 0.73 & 0.74 & 0.82 & 0.78\\
  (Showing work)& XGBoost & 0.77 & 0.59 & 0.67 & 0.69 & 0.83 & 0.75\\
 \hline
\end{tabular}
\end{center}
\caption{Performance Prediction Baseline Models using Ground Truth Equations}
\label{table:mltable}
\end{table}

\section{Related Work}
\label{sec:rw}
The goal of this challenge dataset is to develop methods to introspect a given MWP in order to identify how an LLM (in this case ChatGPT) will perform.  Recent research in this area has examined MWPs can be solved by providing a step-by-step derivation~\cite{gong_continual_2022,ki_generating_2020,kim_ept-x_2022,xia_reasonfuse_2023}.  While these approaches provide insight into potential errors that can lead to incorrect results, this has not been studied in this prior work.  Further, the methods of the aforementioned research are specific to the algorithmic approach.  Work resulting from the use of our challenge dataset could lead to solutions that are agnostic to the underlying MWP solver - as we treat ChatGPT as a black box.  We also note that, if such efforts to introspect MWPs are successful, it would likely complement a line of work dealing with ``chain of thought reasoning'' for LLMs~\cite{wei_chain--thought_nodate,wang_self-consistency_2022} which may inform better ways to generate MWP input into an LLM (e.g., an MWP with fewer additions may be decomposed into smaller problems).  While some of this work also studied LLM performance on Math Word Problems (MWPs), it only looked at how various prompting techniques could improve performance rather than underlying characteristics of the MWP that leads to degraded performance of the LLM.

\section{Future Work}
\label{sec:lim}
Understanding the performance of commercial black-box LLMs will be an important topic as they will likely become widely used for both commercial and research purposes.  Further future directions would also include an examination of ChatGPT performance on datasets other MWPs~\cite{lan_mwptoolkit_2022}, investigating ChatGPT's nondeterminism, and exploring these studies on upcoming commercial LLM's to be released by companies such as Alphabet and Meta.

\begin{acknowledgments}
Some of the authors have been funded by the ASU Fulton Schools of Engineering.
\end{acknowledgments}



\end{document}